\documentclass{article}
\usepackage{ijcai25}
\usepackage{times}
\usepackage{soul}
\usepackage{url}
\usepackage[hidelinks]{hyperref}
\usepackage[utf8]{inputenc}
\usepackage[small]{caption}
\usepackage{graphicx}
\usepackage{amsmath}
\usepackage{amsthm}
\usepackage{booktabs}
\usepackage{xcolor}
\usepackage{multirow}
\usepackage{pifont}
\usepackage{booktabs}

\usepackage{forest}
\usepackage{array}
\usepackage{tikz}
\usetikzlibrary{shadows.blur}

\definecolor{g}{RGB}{220,245,220}
\definecolor{o}{RGB}{250,235,220}

\usepackage[switch]{lineno}

\usepackage[ruled,vlined,linesnumbered]{algorithm2e}
\SetKwInOut{KwIn}{input}
\SetKwInOut{KwOut}{output}
\SetKwFor{Foreach}{\textbf{for each}}{ \textbf{do}}{}
\newcommand{\cmark}{\ding{51}} 
\newcommand{\xmark}{\ding{55}} 
\newcommand{\omark}{$\circ$}   

\usepackage{amssymb}
\usepackage{pifont}
\usepackage{subcaption}
\usepackage{newfloat}
\usepackage{listings}
\usepackage{diagbox}
\usepackage{pifont}   
\usepackage{array}    
\usepackage{graphicx} 

\title{NVIDIA Isaac Sim: Enabling Scalable, GPU-Accelerated Simulation for Robotics}

\author{
Sicong Gao$^1$
\and
Maurice Pagnucco$^1$
\and
Tomasz Bednarz$^2$
\and
Yang Song$^1$
\affiliations
$^1$School of Computer Science and Engineering, The University of New South Wales\\
$^2$NVIDIA USA\\
\emails
\{sicong.gao, morri, yang.song1\}@unsw.edu.au,
tbednarz@nvidia.com
}
\begin{document}

\maketitle

\begin{abstract}
Simulation has become a core infrastructure for robotics research. Unlike previous simulators, NVIDIA Isaac Sim leverages GPU acceleration to enable large-scale parallel training and physics-accurate modeling. Its synthetic data generation pipeline alleviates the scarcity of high-quality training data, supporting data-driven robot learning and large-scale simulation-centric experimentation. However, existing surveys often treat it as one simulator among many, without a systematic analysis of its architectural characteristics, usage patterns, and limitations. This survey reviews Isaac Sim from system and application perspectives, outlining its architecture and comparing it with widely used simulators. We analyze representative studies across five major domains and summarize common usage patterns, particularly in data generation and high-fidelity simulation. We also outline key future directions and challenges, including physics open-world learning, simulation-centric training and practical usability constraints.
\end{abstract}


\begin{figure*}[t]
  \centering
  \includegraphics[width=\linewidth]{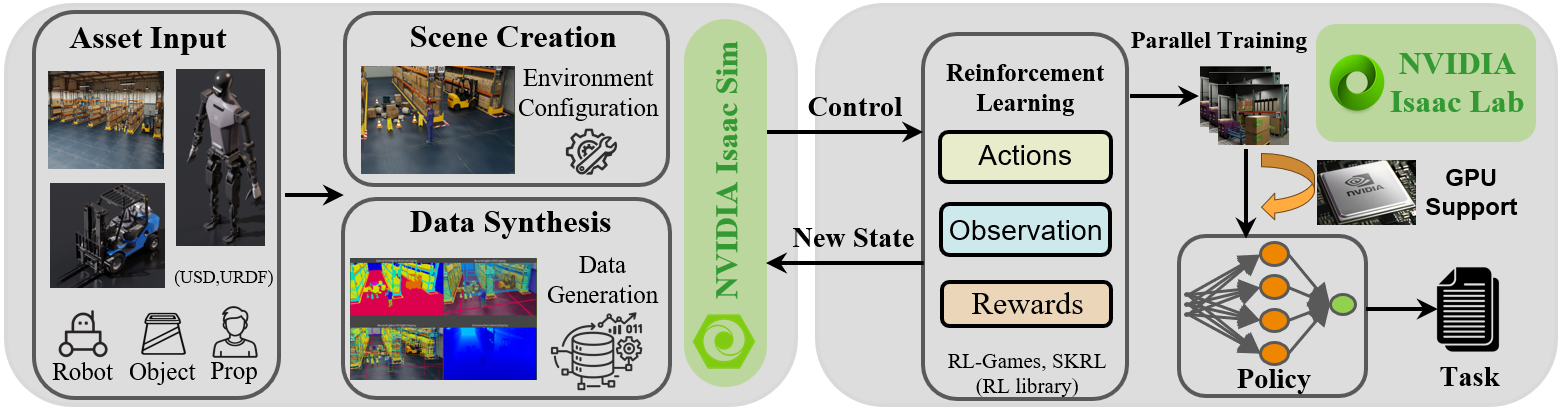}
  \caption{Overall pipeline of reinforcement learning training by integrating NVIDIA Isaac Sim with NVIDIA Isaac Lab. Isaac Sim is used to build high-fidelity simulation scenes and generate synthetic training data, while Isaac Lab provides vectorized environments and RL interfaces (observations, actions, rewards) for GPU-parallel policy optimization.}
  \label{fig:isaacsim_isaaclab_pipeline}
\end{figure*}

\section{Introduction}
Simulation has long been a core tool in robotics and embodied AI, providing a controlled, repeatable environment for developing and evaluating algorithms before deployment in the physical world. As robotic systems increasingly integrate perception, decision-making, and physical interaction, the role of simulation has expanded beyond control and motion verification to encompass perception learning, multimodal data generation, digital twin construction, and system-level evaluation. This evolution places new demands on simulation platforms, which must support not only accurate physical dynamics but also high-fidelity sensing, scalable execution, and tight integration with learning frameworks.

Over the past decade, simulation platforms have been adopted in the robotics community, such as Gazebo~\cite{koenig2004design}, MuJoCo~\cite{todorov2012mujoco}, PyBullet~\cite{coumans2016pybullet}, and Webots~\cite{michel2004cyberbotics}. These simulators have enabled progress in areas such as motion planning, reinforcement learning, navigation, and control. However, many of them are designed with a narrow focus, often emphasizing control-oriented simulation, lightweight physics modeling, or domain-specific scenarios. Consequently, their support for photorealistic rendering, large-scale synthetic data generation, and physical interaction remains limited. As a result, researchers rely on multiple tools or custom pipelines to bridge the gap between physics simulation, sensor modeling, and data-driven learning.

Against this backdrop, NVIDIA Isaac Sim\footnote{\label{fn:isaacsim}\url{https://docs.isaacsim.omniverse.nvidia.com/latest/index.html}} has emerged as a comprehensive simulation platform that addresses these limitations by unifying high-fidelity physics, photorealistic rendering, scalable assets, and native support for robotic learning frameworks. Built on the NVIDIA Omniverse ecosystem, Isaac Sim combines GPU-accelerated PhysX dynamics, RTX-based rendering, and Universal Scene Description (USD)-based scene representation, enabling large-scale simulation of complex environments, diverse sensors, and rich physical interactions. Beyond conventional robotics simulation, it provides systematic support for synthetic data generation, digital twin construction, and large-scale parallel training, making it increasingly relevant to contemporary research in robotic embodied AI.

In recent years, Isaac Sim has been adopted across a wide range of application domains, including industrial, healthcare, and household. It is used not only as a testbed for algorithmic evaluation, but also as a data-centric simulation platform that supports the systematic generation of perception data and the analysis of long-horizon, system-level robotic behaviors that are difficult to study in real-world settings. Despite its growing impact, existing surveys on robotic simulation typically treat Isaac Sim as one platform among many, often focusing on general simulator comparisons or specific application domains \cite{wong2025survey,long2025survey,kargar2024emerging}. To the best of our knowledge, there has been no systematic and dedicated review of Isaac Sim that covers its architecture, application patterns, and limitations.

This survey presents the first comprehensive study of NVIDIA Isaac Sim. Rather than treating it as an isolated tool, we view Isaac Sim as an integrated simulation ecosystem and examine how its core components support the different stages of robotic research, from environment construction and data generation to learning and sim-to-real transfer. We begin by reviewing its architectural design and situating Isaac Sim in the context of other widely used simulation platforms. We then analyze its role in synthetic data generation, with particular attention to programmable scene construction and trajectory learning. On this basis, we survey representative studies across five major domains, including industrial and manufacturing, healthcare and surgical, embodied AI, foundation models, and benchmarks, to illustrate how simulation is leveraged to meet domain-specific task requirements. We also discuss future research directions and challenges. 

By synthesizing recent research, this survey aims to provide researchers and practitioners with a structured understanding of what Isaac Sim enables, how it is used in practice, and its limitations. We hope this work will serve as a reference for selecting appropriate simulation tools, designing simulation-based research pipelines, and identifying open challenges at the intersection of simulation, learning, and embodied intelligence.

\section{Isaac Sim Architecture}
Isaac Sim is designed as a unified simulation infrastructure for robotics and embodied AI, integrating high-fidelity physics, photorealistic rendering, scalable asset management, and seamless interfaces to learning and control frameworks. In this section, we position Isaac Sim in comparison with other simulators (\S\ref{subsec:platform}) and review its core architectural components, including the asset system (\S\ref{subsec:asset}), IsaacLab (\S\ref{subsec:isaaclab}), and digital-twin capabilities (\S\ref{subsec:Digital}).
\subsection{Simulation Platform}
\label{subsec:platform}
Various simulation platforms have been developed for robotics, reinforcement learning, and data-driven perception.
Table~\ref{tab:simulator_feature_comparison} compares representative simulators from multiple perspectives. Among the compared platforms, Isaac Sim stands out as the most comprehensive solution. It natively supports ray tracing and parallel rendering, enabling high-fidelity visual simulation through an RTX-based pipeline.

In contrast, traditional robotics simulators such as Gazebo, MuJoCo, and PyBullet mainly focus on rigid-body dynamics and control-oriented simulation.
While lightweight and efficient, they lack native support for advanced rendering, synthetic data pipelines, and complex physical dynamics, limiting their use to algorithmic evaluation rather than high-fidelity perception or industrial digital twin scenarios~\cite{wong2025survey}. Webots and CoppeliaSim~\cite{rohmer2013v} provide richer sensor support but are largely restricted to rigid-body modeling, with cloth behavior only approximated. CARLA~\cite{dosovitskiy2017carla} is tailored for autonomous driving, offering strong support for vehicle dynamics and sensor simulation, but its applicability is largely confined to traffic environments. Habitat~\cite{puig2023habitat} emphasizes efficient embodied AI and navigation with parallel simulation and reinforcement learning integration, yet does not natively support advanced rendering or complex physical interactions~\cite{long2025survey}. Unity~\cite{juliani2018unity} offers flexible rendering and a mature ecosystem but requires additional engineering to achieve high physical fidelity and robotics middleware integration~\cite{flores2025ros}. In contrast, Isaac Sim provides broader support for physical dynamics and tight integration with ROS and digital twin workflows, making it well-suited for industrial robotics and vision applications.

\begin{table*}[t]
\centering
\Large
\resizebox{\linewidth}{!}{%
\begin{tabular}{lccccccccc}
\toprule
\textbf{Simulator} &
\textbf{Physics Engine} &
\textbf{Ray} &
\textbf{Parallel} &
\textbf{Sensors} &
\textbf{RL} &
\textbf{SynData} &
\textbf{ROS1/2} &
\textbf{Digital Twin/Industry} &
\textbf{Dynamics} \\
& 
& \textbf{Tracing} 
& \textbf{Rendering} 
& \textbf{Stack} 
& \textbf{Integration} 
& \textbf{Pipeline} 
& \textbf{Integration} 
& 
& \\
\midrule

Isaac Sim\textsuperscript{\ref{fn:isaacsim}} &
PhysX (GPU) &
\cmark &
\cmark &
\cmark &
\cmark &
\cmark &
\cmark &
\cmark &
R;D;C;F \\

Gazebo &
ODE / DART / Bullet / Simbody &
\xmark &
\xmark &
\cmark &
\omark &
\xmark &
\cmark &
\omark &
R \\

MuJoCo &
MuJoCo &
\xmark &
\xmark &
\omark &
\cmark &
\xmark &
\omark &
\xmark &
R;D;C \\

PyBullet &
Bullet &
\xmark &
\xmark &
\omark &
\cmark &
\xmark &
\omark &
\xmark &
R;D;C \\

Webots &
ODE &
\xmark &
\xmark &
\cmark &
\omark &
\xmark &
\cmark &
\omark &
R;C$^{\circ}$ \\

CoppeliaSim &
Bullet / ODE / Vortex / Newton &
\omark &
\xmark &
\cmark &
\omark &
\xmark &
\cmark &
\omark &
R;C$^{\circ}$ \\

CARLA &
PhysX &
\omark &
\omark &
\cmark &
\omark &
\omark &
\omark &
\xmark &
R \\

Habitat &
Bullet &
\xmark &
\omark &
\omark &
\cmark &
\omark &
\xmark &
\xmark &
R \\

Unity &
PhysX &
\omark &
\omark &
\omark &
\cmark &
\cmark &
\omark &
\omark &
R;C \\

\bottomrule
\end{tabular}%
}
\caption{Functional comparison of popular simulation platforms, covering rendering, sensor simulation, reinforcement learning (RL) support, synthetic data generation, ROS/ROS~2 integration, digital-twin capabilities, and physical dynamics (R: rigid, D: deformable, C: cloth, F: fluid). \cmark~denotes strong native support; \omark~indicates partial or plugin-based support; \xmark~denotes limited or no support.}
\label{tab:simulator_feature_comparison}
\end{table*}


\subsection{Asset System}
\label{subsec:asset}
Isaac Sim provides a comprehensive set of USD-based assets for constructing simulated environments and robotic systems.
These assets include up to 50 robot models (e.g., articulated manipulators, mobile robots, and mobile manipulators), more than 20 sensor types (e.g., RGB/depth cameras, LiDAR, and IMU), and human characters (e.g., Worker, Police, Doctor). Environment assets span from simple indoor scenes to warehouse and outdoor layouts, supporting diverse testbeds for navigation, manipulation, and interaction tasks. 
The availability of physics-compatible assets has supported recent research efforts.
For example, Orbit \cite{mittal2023orbit} leverages Isaac Sim assets to construct a suite of robotic environments for learning control and interaction policies, including tasks such as cloth folding and cabinet opening with articulated robots.
Similarly, InteriorAgent \cite{InteriorAgent2025} provides indoor scenes in USD format, which have been used to evaluate embodied AI tasks such as navigation and layout understanding with high-fidelity visual detail.

\subsection{IsaacLab}
\label{subsec:isaaclab}
NVIDIA IsaacLab\footnote{\label{fn:isaaclab}\url{https://isaac-sim.github.io/IsaacLab/main/index.html}} is a reinforcement learning framework built on Isaac Sim that integrates libraries such as RL-Games, SKRL, and RSL-RL for large-scale robot learning. As shown in Fig.~\ref{fig:isaacsim_isaaclab_pipeline}, NVIDIA Isaac Sim and Isaac Lab are combined to form an end-to-end reinforcement learning training pipeline.
It provides standardized interfaces for environment and control definition, and supports GPU-based parallel execution across many environments.
By coupling reinforcement learning with Isaac Sim’s physics engine, IsaacLab enables efficient training under physically consistent simulation.

Peterson et al.~\shortcite{peterson2025framework} propose a heterogeneous adversarial multi-agent reinforcement learning framework for scalable training under high-fidelity physics.
Similarly, MARLadona.~\cite{li2025marladona} is a multi-agent soccer environment that enables thousands of parallel simulations via GPU acceleration. This scalability is supported by IsaacLab’s vectorized environments and its tight integration with the Isaac physics engine. Wheeled Lab \cite{han2025wheeled} integrates wheeled robots with IsaacLab to support large-scale reinforcement learning for control and perception-driven navigation. Using IsaacLab’s realistic simulation and modular environments, policies are trained for tasks such as controlled drifting, elevation traversal, and vision-based navigation.

\subsection{Digital Twin}
\label{subsec:Digital}
Digital twin technology uses virtual models to mirror real-world systems for monitoring, prediction, and optimization.
By integrating sensing, data, and simulation, it synchronizes virtual and physical entities to support decision-making. Isaac Sim provides high-fidelity digital twin modules that enable rapid construction of digital twin systems, including warehouse layout builders, conveyor tools, and static warehouse assets, to create complete logistics or industrial scenarios for state monitoring, scheduling optimization, and control policy validation.

Horsgaard et al.~\shortcite{horsgaard2025exploring} construct a warehouse environment and integrate a Grab™ robot to enable package picking with a robotic arm and gripper. Compared to earlier Webots-based simulations, their approach addresses limitations related to unrealistic environments and slow execution speed. Building on similar warehouse sorting scenarios, Sand et al.~\shortcite{sandutilizing} use the built-in conveyor system together with monocular vision and reinforcement learning to achieve real-time grasping and sorting. In contrast, Nambiar et al.~\shortcite{nambiar2024automation} apply Isaac Cortex to microscopic slide sorting by transferring ROS commands to a real ABB YuMi robot. While this approach enables synchronization between the simulation and the real system, it lacks feedback from the physical environment, as perception relies on predefined spreadsheet data rather than real sensors, which limits realism and adaptability.


\section{Synthetic Data and Perception}
High-quality data is essential for data-driven robotics, yet real-world collection is limited by cost, safety, and scalability. Simulation-based synthetic data generation has therefore become a key paradigm for controllable and scalable learning. This section reviews how Isaac Sim enables synthetic data generation across perception (\S\ref{subsec:Perception}), actor-centric simulation (\S\ref{subsec:Actor}), and robotic trajectory modeling (\S\ref{subsec:Trajectory}) through high-fidelity simulation, programmable scene construction, and large-scale randomization.

\subsection{Synthetic Data in Robotic Perception}
\label{subsec:Perception}
Robotic perception models require large-scale annotated data, yet real-world collection is costly and constrained by annotation effort, limited long-tail coverage, and safety concerns. Synthetic data generation mitigates these issues by providing scalable data with accurate labels and controllable distributions that are difficult to obtain in real environments.

In this context, Replicator\footnote{\label{fn:replicator}\url{https://docs.isaacsim.omniverse.nvidia.com/latest/replicator_tutorials/index.html}}, an extension of Isaac Sim, provides a programmable framework for synthetic data generation in perception learning, supporting semantic annotation in USD-based scenes. For instance, Elia et al. \shortcite{bonetto2023grade} built a multi-robot aerial–ground robotic platform for SLAM. Once configured, these scenes can be rendered with GPU acceleration to produce richly annotated multimodal datasets, including bounding boxes, segmentation masks, and depth maps, etc. Ng et al. \shortcite{ng2023syntable} proposed SynTable, which uses Replicator to generate complex 3D tabletop scenes with diverse object meshes, materials, textures, lighting conditions, and backgrounds. Their pipeline targets amodal instance segmentation for previously unseen objects and provides supervision signals that are difficult to obtain from real-world data.

Beyond static scene generation, Replicator also supports large-scale scene and sensor randomization. Ritvik et al. \shortcite{singh2024synthetica} leveraged Replicator’s asset library and procedural randomization to generate diverse indoor environments for object detection. By programmatically constructing room layouts, placing objects, randomizing camera trajectories, and introducing distractors to increase occlusion diversity, their approach reduces manual annotation effort and improves robustness under complex visual conditions.


\usetikzlibrary{arrows.meta, shadows.blur}

\forestset{
  domainTag/.style={
    draw=black!55,
    line width=0.45pt,
    rounded corners=3pt,
    inner sep=2mm,
    fill=#1!8!white,
    text=black,
    minimum height=8mm
  },
  rootTag/.style={
    draw=black!55,
    line width=0.45pt,
    rounded corners=3pt,
    inner sep=2mm,
    fill=cyan!25,
    rotate=90,
    text=black,
    minimum height=10mm,
    font=\normalsize
  },
  citeCard/.style={
    draw=black!45,
    line width=0.45pt,
    rounded corners=3pt,
    inner sep=2mm,
    fill=#1!4!white,
    blur shadow={
      shadow xshift=0.8pt,
      shadow yshift=-0.8pt,
      shadow blur steps=6,
      shadow opacity=0.12
    }
  }
}

\begin{figure*}[!t]
\centering

\begin{forest}
for tree={
  grow'=east,
  parent anchor=east,
  child anchor=west,
  anchor=west,
  align=left,
  edge={-{Latex}, line width=0.6pt},
  l sep=12pt,
  s sep=8pt,
  font=\small,
},
where level=0{parent anchor=south}{}
[
  {Domain Applications},
  rootTag,
  minimum width=40mm,
  xshift=-1.8cm
  [
    {Industrial \&\\Manufacturing},
    domainTag=teal,
    minimum width=32mm
    [
      {\parbox[t]{125mm}{
      Kagami et al.~\shortcite{kagami2025multi},
      Imran et al.~\shortcite{imran2024decentralized},
      Haug et al.~\shortcite{haug2025vision},
      Chen et al.~\shortcite{chen2025task},
      Koprov et al.~\shortcite{koprov2025industrial},
      Nguyen et al.~\shortcite{nguyen2025efficient},
      Monnet et al.~\shortcite{monnet2024investigating},
      Jeong et al.~\shortcite{jeong2022digital}
      }},
      citeCard=teal
    ]
  ]
  [
    {Healthcare \&\\Surgery},
    domainTag=green,
    minimum width=32mm
    [
      {\parbox[t]{125mm}{
      Gao et al.~\shortcite{gao2026reinforcementlearningfollowtheleaderrobotic},
      Kim et al.~\shortcite{kim2025surgical},
      ORBIT-Surgical~\cite{yu2024orbit},
      Hydock et al.~\shortcite{hydock2023generation},
      dARt Vinci~\cite{liu2025dart},
      SUFIA~\cite{moghani2024sufia},
      SUFIA-BC~\cite{moghani2025sufia}
      }},
      citeCard=green
    ]
  ]
  [
    {Embodied AI},
    domainTag=blue,
    minimum width=32mm
    [
      {\parbox[t]{125mm}{
      AdaVLN~\cite{loh2024adavln},
      ReKep~\cite{huang2024rekep},
      Bonyani et al.~\shortcite{bonyani2025embodied},
      TacEx~\cite{nguyen2024tacex},
      GRADE~\cite{bonetto2023grade},
      PR2~\cite{liu2025pr2},
      ManiSkill3~\cite{tao2024maniskill3}
      }},
      citeCard=blue
    ]
  ]
  [
    {Foundation\\Models},
    domainTag=orange,
    minimum width=32mm
    [
      {\parbox[t]{125mm}{
      NVIDIA GR00T~\cite{bjorck2025gr00t},
      IsaacLabEvalTasks~\cite{IsaacLabEvalTasks},
      Park et al.~\shortcite{park2025modality},
      Cosmos~\cite{agarwal2025cosmos}
      }},
      citeCard=orange
    ]
  ]
  [
    {Platform \&\\Benchmark},
    domainTag=purple,
    minimum width=32mm
    [
      {\parbox[t]{125mm}{
      InfiniteWorld~\cite{ren2024infiniteworld},
      AgentWorld~\cite{zhang2025agentworld},
      RealMirror~\cite{tai2025realmirror},
      BEHAVIOR-1K~\cite{li2023behavior},
      Kitchen-R~\cite{kachaev2025mind},
      Zhou et al.~\shortcite{zhou2024towards},
      Lin et al.~\shortcite{lin2025vlnverse}
      }},
      citeCard=purple
    ]
  ]
]
\end{forest}

\caption{Major domain applications of Isaac Sim and representative studies in each domain.}
\label{fig:domain_applications}
\end{figure*}
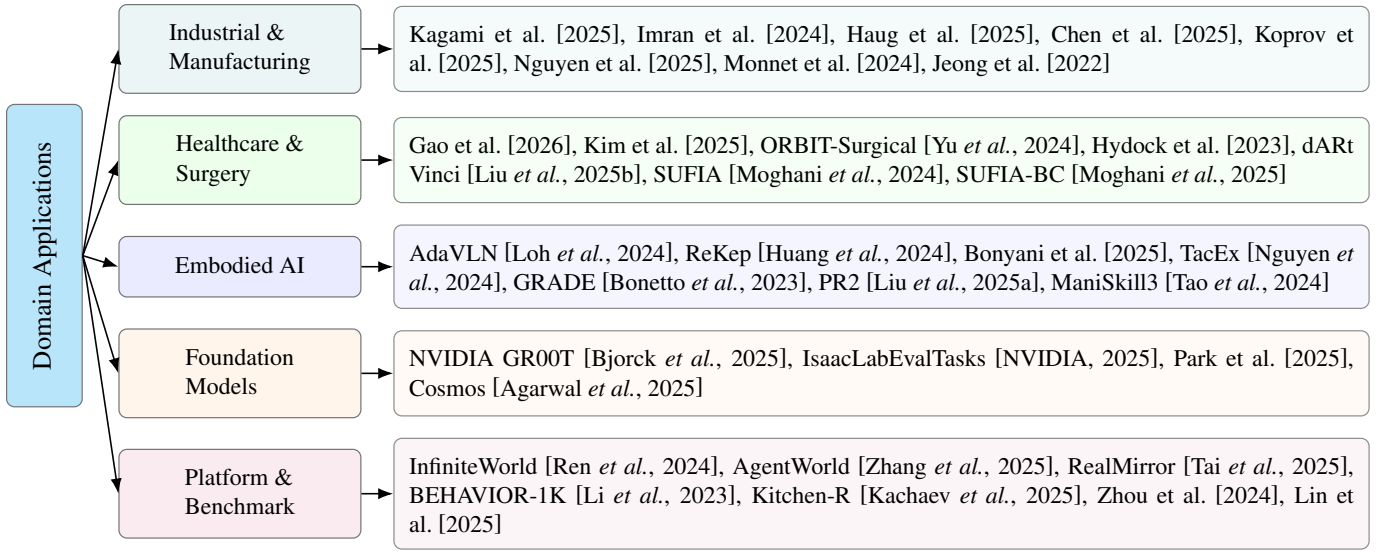

\subsection{Synthetic Data in Actor Simulation}
\label{subsec:Actor}
Real-world data collection is often limited by scalability, cost, and the difficulty of capturing rare or hazardous scenarios such as accidents or near-miss events. Simulation enables these long-tail events to be generated programmatically. Isaac Sim supports actor and event-oriented synthetic data workflows, enabling the generation of dynamic, richly annotated sequences by simulating in realistic environments. Maric et al. \shortcite{maric2022large} constructed a large-scale dataset for indoor fire detection and segmentation, systematically covering diverse indoor environments and challenging lighting conditions to address the difficulty of collecting real-world fire and other hazardous scenarios. OmniGibson \cite{li2023behavior} is a large-scale interactive indoor environment for embodied AI. Developing robot-environment interaction processes through home-based scenarios. Andreas et al. \shortcite{andreas2023deformable} constructed a pipeline of conveyor–fish interactions in a fish factory. This workflow enables systematic testing of fish sorting and grasping strategies while allowing safe, repeatable simulation of complex physical interactions that would be costly and difficult to scale in real-world settings.

\subsection{Synthetic Data in Robotic Trajectory}
\label{subsec:Trajectory}
Real-world trajectory collection for mobile robots is costly and often constrained by safety and scalability. Simulation enables systematic synthesis of trajectory data under controlled conditions. Isaac Sim addresses this need through the MobilityGen pipeline, which captures time-series robot trajectories and multimodal sensor data during simulated navigation, enabling scalable datasets for tasks such as navigation, SLAM, and motion prediction. RoboMIND \cite{wu2024robomind} contains 107k manipulation trajectories across multiple robot embodiments, each annotated with multi-view RGB-D observations, robot states, and task descriptions. By aligning simulation with physical setups, Isaac Sim enables low-cost demonstration augmentation and systematic evaluation of manipulation policies across diverse tasks and robot embodiments. Salimpour et al. \shortcite{salimpour2025sim} leveraged MobilityGen to train reinforcement-learning navigation and obstacle-avoidance policies in physics-enabled simulation environments. The policies were subsequently transferred to Gazebo and to real ROS 2 robots in a zero-shot manner. Kitchen-R \cite{kachaev2025mind} enables a large-scale collection of robotic trajectories for mobile manipulation tasks in Isaac Sim. It records navigation and manipulation trajectories while a mobile manipulator executes multi-step language instructions in a simulated kitchen, logging time-series robot states, gripper commands, and synchronized multi-view sensor observations in ROS-compatible formats.

\section{Domain Applications}
Isaac Sim is widely used across domains where high-fidelity simulation is essential for development and evaluation. This section reviews representative applications built on Isaac Sim, including industrial and manufacturing robotics (\S\ref{subsec:industrial}), healthcare and surgery (\S\ref{subsec:healthcare}), embodied AI (\S\ref{subsec:embodied}), foundation models (\S\ref{subsec:foundation}), and platform and benchmark systems (\S\ref{subsec:benchmark}), highlighting how task-specific requirements are addressed through simulation. The categorization of application domains and representative studies is illustrated in Fig.~\ref{fig:domain_applications}.

\subsection{Industrial and Manufacturing}
\label{subsec:industrial}
Industrial and manufacturing environments demand robotic systems that are safe, efficient, and robust under complex layouts and human–robot coexistence. Isaac Sim can realistically model warehouse layouts, robot kinematics, sensors, and human activities, making it well-suited for developing and evaluating automated guided vehicles (AGVs) and autonomous mobile robots (AMRs). Kagami et al. \shortcite{kagami2025multi} investigated multi-AGV coordination for robot-to-parts picking, analyzing shelf retrieval and task scheduling behaviors. Imran et al. \shortcite{imran2024decentralized} evaluated warehouse layouts and dispatch strategies, highlighting Isaac Sim’s value for system-level AGV analysis. Haug et al. \shortcite{haug2025vision} studied AMR safety in human–robot shared spaces. Isaac Sim also enables the industrial metaverse by linking physical assets, digital twins, and intelligent decision systems. Chen et al. \shortcite{chen2025task} view the industrial metaverse as a data-driven ecosystem supporting workflow simulation, such as sorting and production. Koprov et al. \shortcite{koprov2025industrial} further demonstrate that integrating Isaac Sim with industrial IoT pipelines enables real-time synchronization and motion simulation for industrial metaverse applications. In addition, Isaac Sim can be applied to industrial inspection by programmatically simulating surface defects under diverse viewpoints and lighting conditions, thereby supporting robust training and evaluation of inspection models for factory components \cite{nguyen2025efficient,monnet2024investigating,jeong2022digital}.

\subsection{Healthcare and Surgery}
\label{subsec:healthcare}
Due to tissue deformability and constrained workspaces, surgical robots require high perception accuracy, reliable contact-rich manipulation, and strict safety guarantees. Gao et al. \cite{gao2026reinforcementlearningfollowtheleaderrobotic} leverage Isaac Sim to generate large-scale synthetic endoscopic datasets for depth estimation and navigation. Beyond perception, Isaac Sim also supports physically accurate simulation of surgical manipulation. Kim et al.~\shortcite{kim2025surgical} establish a robot-assisted suturing environment, where Isaac Sim’s ROS interfaces are integrated with haptic teleoperation to enable interactive training and reproducible benchmarking of suturing skills. At the system level, ORBIT-Surgical~\cite{yu2024orbit} builds an open simulation benchmark on Isaac Sim for training and evaluating reinforcement learning and imitation learning policies on fundamental surgical subtasks, such as needle extraction, handover, and tissue manipulation, while supporting large-scale parallel simulation and synthetic perception data generation. 

To further reduce reliance on physical hardware and expert demonstrations, Hydock et al. \cite{hydock2023generation} use Isaac Sim to render surgical instruments in randomized environments, automatically generating precisely annotated medical images for training object detection and semantic segmentation models. dARt Vinci \cite{liu2025dart} introduces an egocentric data collection pipeline that combines Isaac Sim with Augmented Reality-based interfaces, allowing users to collect surgical robot learning data using only a head-mounted display and a standard PC. Building on this paradigm, SUFIA \cite{moghani2024sufia} and SUFIA-BC \cite{moghani2025sufia} integrate language-guided planning, anatomically realistic digital twins, and large-scale synthetic demonstrations within Isaac Sim, enabling systematic evaluation of behavior cloning and language-conditioned autonomy for surgical assistants.



\subsection{Embodied AI}
\label{subsec:embodied}
Embodied artificial intelligence enables perception, reasoning, and action through physical interaction. Isaac Sim integrates photorealistic rendering and multi-sensor modeling, making it well-suited for vision–language embodied learning in physically grounded environments. Recent studies leverage Isaac Sim to investigate instruction-driven embodied navigation in unstructured settings. AdaVLN \cite{loh2024adavln} extends vision–language navigation to continuous indoor environments with moving humans, and introduces a freeze-time mechanism to ensure fair and reproducible evaluation under dynamic obstacles. ReKep \cite{huang2024rekep} models tasks as spatio-temporal relational keypoint constraints, addressing the challenge of converting free-form language instructions into executable embodied behaviors and enabling real-time robot action generation. Bonyani et al. \shortcite{bonyani2025embodied} integrate LiDAR and depth sensing into Isaac Sim to support instruction-aware planning and collision-free navigation, demonstrating Isaac Sim's advantage in unifying realistic sensing, dynamics, and instruction grounding.

High-fidelity physical interaction simulation is a fundamental component of embodied intelligence. Isaac Sim provides core support for complex embodied interaction by unifying physics-consistent simulation with photorealistic sensor rendering. TacEx \cite{nguyen2024tacex} enables high-fidelity contact modeling for GelSight visuotactile sensors, GRADE \cite{bonetto2023grade} supports the study of active SLAM and multi-robot perception in realistic dynamic environments, and PR2~\cite{liu2025pr2} integrates accurate dynamics with indoor and outdoor scenes to study the joint embodiment of locomotion, manipulation, and perception in humanoid robots. At scale, ManiSkill3~\cite{tao2024maniskill3} extends the GPU-based physics simulation paradigm of Isaac simulators, enabling efficient large-scale embodied reinforcement and imitation learning through parallel training.

\subsection{Foundation Models}
\label{subsec:foundation}
Robotic foundation models unify perception, language, and action, relying on high-fidelity simulation such as Isaac Sim for scalable training and evaluation.
NVIDIA GR00T~\cite{bjorck2025gr00t} is a foundation model for generalist robots based on a unified vision–language–action formulation. It relies on Isaac Sim for large-scale, high-fidelity simulation to enable reproducible data generation and improved generalization with fewer real-world data samples. It defines two industrial manipulation tasks \cite{IsaacLabEvalTasks} in Isaac Sim: nut pouring and exhaust pipe sorting, each requiring multi-step bimanual humanoid actions. In addition, Park et al. \shortcite{park2025modality} build on GR00T’s diffusion policy and use Isaac Sim simulations as well as physical interaction modeling to improve stability and generalization in grasp-and-place tasks.

Cosmos \cite{agarwal2025cosmos} is a world model proposed by NVIDIA for Physical AI, enabling the generation of physically consistent videos and scene trajectories for robot learning. It complements Isaac Sim, which provides high-fidelity simulation and digital twins, while Cosmos extends data coverage through generative world modeling to capture long-horizon dynamics, rare events, and complex interactions, and to generate realistic scene variations from simulation priors.

\subsection{Platform and Benchmark}
\label{subsec:benchmark}
Many studies regard Isaac Sim as an extensible simulation platform adaptable to task-specific scenarios. 
InfiniteWorld~\cite{ren2024infiniteworld} focuses on open-world vision–language interaction, emphasizing long-horizon interaction under partially observable environments. It aims to evaluate how agents explore and act in large-scale environments with continuously generated scenes. AgentWorld \cite{zhang2025agentworld}  focuses on household mobile manipulation, with its core contribution being the integration of procedural scene construction and mobile teleoperation to collect long-horizon manipulation trajectories across diverse home layouts. RealMirror \cite{tai2025realmirror} is designed for vision–language–action learning in humanoid robots, with a particular emphasis on evaluation and sim-to-real transfer.

Benchmarks evaluate an agent’s ability to perceive, plan, and act under physical constraints. Isaac Sim enables assessment of physical feasibility, efficiency, and robustness beyond task success rates. BEHAVIOR-1K~\cite{li2023behavior} evaluates performance on 1,000 everyday household activities, while Kitchen-R~\cite{kachaev2025mind} focuses on language-guided mobile manipulation in a realistic kitchen digital twin and explicitly separates task planning and low-level control. The AI-CPS \cite{zhou2024towards} benchmark evaluates industrial manipulation tasks, including peg-in-hole insertion, stacking, door opening, and cloth placement. VLNVerse~\cite{lin2025vlnverse} evaluates vision–language navigation under continuous control and full embodiment.

\section{Future Directions and Challenges}
\paragraph{Physically Executable Open-World Learning}
Recent open-world research has increasingly shifted toward generative world models that emphasize large-scale environment synthesis and semantic diversity. While these approaches can produce visually rich environments, they often lack physically consistent interaction, including reliable collision handling, contact dynamics, and long-horizon physical feasibility. Lu et al. \shortcite{kang2024far} note that robots trained in open-world settings tend to rely on case-based generalization rather than abstract physical rules, while Mao et al. \shortcite{mao2025robot} show that directly applying video generation models to robot control neglects physical constraints.
These limitations restrict the deployment of generative open-world models in real robotic systems. In contrast, Isaac Sim provides a physics-consistent simulation backbone that complements generative models by enforcing physical validity and executability. A promising direction is hybrid open-world systems that integrate generative scene synthesis with high-fidelity physics simulation. Generative models offer diverse and evolving environments, while Isaac Sim enforces physically grounded interaction, enabling scalable training and evaluation in executable open worlds. Its GPU-accelerated physics further enables the incorporation of physical constraints into generative model training, bridging visual diversity and physical realism.

\paragraph{Toward Simulation-Centric Training of Robotic Foundation Models}
Robotic foundation models require large-scale, multimodal, and physically realistic data to jointly learn perception, language, and action; however, real-world data collection is costly, slow, and constrained by safety and scalability. Isaac Sim provides a cost-effective alternative as a unified simulation-based training platform, where high-fidelity physical modeling helps reduce the sim-to-real gap and enables more transferable representations and policies.
Its native support for GPU acceleration and large-scale parallel simulation further improves training efficiency, allowing foundation models to scale across diverse tasks, instances, and scenarios, and to explore large architectures and long-horizon behaviors within practical time budgets \cite{ahmed2024systemic}. Moreover, the Replicator pipeline enables a data-centric training paradigm by generating large-scale, richly annotated datasets covering visual perception, robot trajectories, human–robot interaction, and multimodal sensor streams, effectively capturing diversity and long-tail scenarios \cite{salimpour2025sim}.
Beyond training, foundation models can be deployed within Isaac Sim for planning, decision-making, and interaction studies, supporting controlled system-level evaluation and safety analysis. This bidirectional integration positions Isaac Sim as both a training platform and a testbed for studying foundation model behavior under physically grounded interaction conditions.

\paragraph{Generation–Simulation Hybrid Approaches for Sim-to-Real Transfer}
One long-standing challenge in robotics is the sim-to-real gap, particularly caused by visual discrepancies between simulated environments and the real world, which hinder reliable policy transfer. Looking ahead, integrating generative world models, such as Cosmos \cite{agarwal2025cosmos}, into physics-based simulation pipelines enabled by Isaac Sim represents a promising approach to address this issue. By converting simulated scenes into realistic digital twins, Isaac Sim can help reduce visual distribution mismatches while preserving consistent physical interaction.
For example, DSR \cite{cordero2025dsr} uses Isaac Sim for large-scale interaction learning in simulation, while Cosmos enriches visual and semantic diversity through generative modeling. Such hybrid simulation–generation approaches may significantly reduce dependence on real-world data collection and alleviate sim-to-real performance degradation caused by visual inconsistencies, particularly in complex robotic navigation scenarios.

\paragraph{Cloud-Native Collaborative Simulation Worlds}
Another promising direction is to extend Isaac Sim toward cloud-native, persistent, and collaborative simulation worlds. In such environments, simulations remain continuously online and are jointly constructed and interacted with by multiple users, robots, and learning agents. These shared simulation worlds resemble small virtual ecosystems, where changes introduced by one participant can affect others in real time, enabling collective experimentation, long-horizon interaction, and continual data accumulation.
Isaac Sim is well positioned to support this direction due to its system-level design. Built on NVIDIA Omniverse, it natively supports distributed rendering, asset streaming, and real-time synchronization, enabling multi-user and multi-agent interaction within shared virtual spaces. In addition, its GPU-accelerated physics engine allows simulation to scale across many environments and agents, making persistent worlds practical on cloud infrastructure with dynamic resource allocation. The use of standardized scene representations and modular simulation components enables environments to be updated, versioned, and shared incrementally across users without requiring entire scenes to be rebuilt.

\paragraph{Challenge: Usability and Learning Curve of Isaac Sim}
Despite its strong performance in physics fidelity and large-scale simulation, the usability and learning curve of Isaac Sim remain practical challenges for broader adoption. Compared with lighter alternatives such as Webots or CoppeliaSim, Isaac Sim incurs substantially higher computational and memory overhead, which increases setup complexity and development costs, particularly for users with limited resources or experience \cite{wong2025survey}. In addition, its reliance on high-end hardware and complex configuration pipelines may pose barriers for smaller research teams, even though it offers advanced GPU acceleration and photorealistic rendering \cite{flores2025ros}. More broadly, high-fidelity simulation platforms such as Isaac Sim often require nontrivial configuration and specialized expertise to fully exploit their capabilities, highlighting a trade-off between realism and ease of use \cite{zafra2025survey}. Addressing these usability challenges through improved abstractions, streamlined workflows, and clearer configuration standards will be critical for enabling wider and more effective use of Isaac Sim in both research and practice.

\section{Conclusions}
This survey reviews recent studies that adopt NVIDIA Isaac Sim across different research domains. The literature indicates that Isaac Sim is used not only as a simulation tool, but more importantly as an integrated platform for data generation and for jointly modeling physical dynamics and perception learning. Its design enables researchers to study embodied decision-making and data-driven robotic foundation models under physically grounded conditions, while leveraging GPU-accelerated rendering and large-scale parallel simulation capabilities that are difficult to achieve with many other simulators. The surveyed works suggest a shift from task-specific simulation toward system-level evaluation. Rather than focusing on isolated scenarios, many studies use Isaac Sim to analyze interaction-rich behaviors, long-horizon execution, and other complex settings, aiming to reduce the sim-to-real gap through more realistic simulation environments. At the same time, the high computational cost and steep learning curve remain practical obstacles to its broader adoption.

\clearpage
\bibliographystyle{named}
\bibliography{main}

@mastersthesis{horsgaard2025exploring,
  title={Exploring Real-Time Robot Simulation in {Isaac Sim} with a Digital Twin of a Warehouse Picking Robot},
  author={Horsgaard, Simon Krogh and others},
  type={{B.S.} thesis},
  year={2025},
  school={NTNU}
}

@mastersthesis{andreas2023deformable,
  title={Deformable Body Based Salmon Model-For iterative development and testing of equipment within {Isaac Sim}},
  author={Andreas, Glomsrud},
  year={2023},
  school={NTNU}
}

@article{sandutilizing,
  title={Utilizing Reinforcement Learning and Computer Vision in a {Pick-And-Place} Operation for Sorting Objects in Motion},
  author={SAND, KRISTOFFER and others},
  year={2023}
}

@article{salimpour2025sim,
  title={{Sim-to-Real} Transfer for Mobile Robots with Reinforcement Learning: from {NVIDIA Isaac Sim to Gazebo} and Real {ROS} 2 Robots},
  author={Salimpour, Sahar and others},
  journal={arXiv:2501.02902},
  year={2025}
}

@article{nambiar2024automation,
  title={Automation in unstructured production environments using {Isaac Sim}: A flexible framework for dynamic robot adaptability},
  author={Nambiar, Sanjay and others},
  journal={Procedia CIRP},
  volume={130},
  pages={837--846},
  year={2024},
}

@article{ng2023syntable,
  title={Syntable: A synthetic data generation pipeline for unseen object amodal instance segmentation of cluttered tabletop scenes},
  author={Ng, Zhili and others},
  journal={arXiv:2307.07333},
  year={2023}
}

@article{singh2024synthetica,
  title={Synthetica: Large Scale Synthetic Data for Robot Perception},
  author={Singh, Ritvik and others},
  journal={arXiv:2410.21153},
  year={2024}
}

@article{bonetto2023grade,
  title={{GRADE}: Generating Realistic and Dynamic Environments for robotics research with Isaac Sim},
  author={Bonetto, Elia and others},
  journal={IJRR},
  pages={02783649251346211},
  year={2023}
}

@inproceedings{li2023behavior,
  title={{Behavior-1K}: A benchmark for embodied {AI} with 1,000 everyday activities and realistic simulation},
  author={Li, Chengshu and others},
  booktitle={CoRL},
  pages={80--93},
  year={2023},
  organization={PMLR}
}

@inproceedings{maric2022large,
  title={A large scale dataset for fire detection and segmentation in indoor spaces},
  author={Maric, Petar and others},
  booktitle={ICECCME},
  pages={1--8},
  year={2022},
  organization={IEEE}
}

@article{wu2024robomind,
  title={{Robomind: Benchmark} on multi-embodiment intelligence normative data for robot manipulation},
  author={Wu, Kun and others},
  journal={arXiv:2412.13877},
  year={2024}
}

@article{wong2025survey,
  title={A Survey of Robotic Navigation and Manipulation with Physics Simulators in the Era of Embodied {AI}},
  author={Wong, Lik Hang Kenny and others},
  journal={arXiv:2505.01458},
  year={2025}
}

@article{long2025survey,
  title={A Survey: Learning Embodied Intelligence from Physical Simulators and World Models},
  author={Long, Xiaoxiao and others},
  journal={arXiv:2507.00917},
  year={2025}
}

@inproceedings{koenig2004design,
  title={Design and use paradigms for {Gazebo}, an open-source multi-robot simulator},
  author={Koenig, Nathan and others},
  booktitle={IROS},
  volume={3},
  pages={2149--2154},
  year={2004},
  organization={IEEE}
}

@inproceedings{todorov2012mujoco,
  title={{Mujoco}: A physics engine for model-based control},
  author={Todorov, Emanuel and others},
  booktitle={IROS},
  year={2012},
  organization={IEEE}
}

@misc{coumans2016pybullet,
  title={Pybullet, a {Python} module for physics simulation for games, robotics and machine learning},
  author={Coumans, Erwin and others},
  year={2016}
}

@inproceedings{rohmer2013v,
  title={{V-REP}: A versatile and scalable robot simulation framework},
  author={Rohmer, Eric and others},
  booktitle={IROS},
  pages={1321--1326},
  year={2013},
  organization={IEEE}
}

@inproceedings{dosovitskiy2017carla,
  title={{CARLA}: An open urban driving simulator},
  author={Dosovitskiy, Alexey and others},
  booktitle={Conference on robot learning},
  pages={1--16},
  year={2017},
  organization={PMLR}
}

@article{puig2023habitat,
  title={Habitat 3.0: A co-habitat for humans, avatars and robots},
  author={Puig, Xavier and others},
  journal={arXiv:2310.13724},
  year={2023}
}

@article{michel2004cyberbotics,
  title={Cyberbotics ltd. webots™: professional mobile robot simulation},
  author={Michel, Olivier},
  journal={IJARS},
  volume={1},
  number={1},
  pages={5},
  year={2004},
}

@article{juliani2018unity,
  title={Unity: A general platform for intelligent agents},
  author={Juliani, Arthur and others},
  journal={arXiv:1809.02627},
  year={2018}
}

@article{mittal2023orbit,
  title={Orbit: A unified simulation framework for interactive robot learning environments},
  author={Mittal, Mayank and others},
  journal={IEEE RA-L},
  volume={8},
  number={6},
  pages={3740--3747},
  year={2023},
}

@misc{InteriorAgent2025,
  title        = {InteriorAgent: Interactive {USD} Interior Scenes for {Isaac Sim}-based Simulation},
  author       = {SpatialVerse Research Team and Manycore Tech Inc.},
  year         = {2025},
}

@inproceedings{li2025marladona,
  title={Marladona-towards cooperative team play using multi-agent reinforcement learning},
  author={Li, Zichong and others},
  booktitle={ICRA},
  pages={15014--15020},
  year={2025},
  organization={IEEE}
}

@article{peterson2025framework,
  title={A Framework for Scalable Heterogeneous Multi-Agent Adversarial Reinforcement Learning in {IsaacLab}},
  author={Peterson, Isaac and others},
  journal={arXiv:2510.01264},
  year={2025}
}

@inproceedings{han2025wheeled,
  title={Wheeled {Lab}: Modern Sim2Real for Low-cost, Open-source Wheeled Robotics},
  author={Han, Tyler and others},
  booktitle={RSS},
  year={2025}

}

@inproceedings{kagami2025multi,
  title={Multi-{AGV} Assisted {OPS} in Warehouse Using Digital Twins and Layout Comparison},
  author={Kagami, Haruka and others},
  booktitle={MetaCom},
  pages={205--212},
  year={2025},
  organization={IEEE}
}

@inproceedings{haug2025vision,
  title={Vision-Based Human Awareness Estimation for Enhanced Safety and Efficiency of {AMRs} in Industrial Warehouses},
  author={Haug, Maximilian and others},
  booktitle={ETFA},
  pages={1--4},
  year={2025},
  organization={IEEE}
}

@inproceedings{imran2024decentralized,
  title     = {Decentralized Multi-Robot Shared Perception for Worker Action Inference in Industrial Facilities},
  author    = {Imran, Ali and others},
  booktitle = {ICRA},
  year      = {2024},
}

@article{chen2025task,
  title={Task-Oriented Edge-Assisted Cross-System Design for Real-Time Human-Robot Interaction in Industrial Metaverse},
  author={Chen, Kan and others},
  journal={arXiv:2508.20664},
  year={2025}
}

@article{koprov2025industrial,
  title={Industrial metaverse meets IIoT: Low-code platforms for machine-to-machine and human-to-machine integration},
  author={Koprov, Pavel and others},
  journal={Manufacturing Letters},
  volume={44},
  pages={1254--1265},
  year={2025},
}

@article{nguyen2025efficient,
  title={Efficient synthetic defect on 3D object reconstruction and generation pipeline for digital twins smart factory},
  author={Nguyen, Viet-Hoan and others},
  journal={Sensors},
  year={2025},
}

@article{monnet2024investigating,
  title={Investigating the generation of synthetic data for surface defect detection: A comparative analysis},
  author={Monnet, Josefine and others},
  journal={Procedia CIRP},
  volume={130},
  pages={767--773},
  year={2024},
}

@inproceedings{jeong2022digital,
  title={Digital twin-based cutting tool breakage detection model using synthetic depth map and deep learning},
  author={Jeong, Suhwan and others},
  booktitle={ASPEN},
  year={2022}
}

@inproceedings{kim2025surgical,
  title={Surgical Robotics Environment in {NVIDIA Isaac Sim} for Robot-Assisted Suturing},
  author={Kim, Tae Wan and others},
  booktitle={ISMR},
  pages={192--198},
  year={2025},
  organization={IEEE}
}

@inproceedings{yu2024orbit,
  title={Orbit-surgical: An open-simulation framework for learning surgical augmented dexterity},
  author={Yu, Qinxi and others},
  booktitle={ICRA},
  pages={15509--15516},
  year={2024},
  organization={IEEE}
}

@inproceedings{hydock2023generation,
  title={Generation of synthetic data for medical decision support applications},
  author={Hydock, Kenneth and others},
  booktitle={AIPR},
  pages={1--7},
  year={2023},
  organization={IEEE}
}

@article{liu2025dart,
  title={{dARt Vinci}: Egocentric Data Collection for Surgical Robot Learning at Scale},
  author={Liu, Yihao and others},
  journal={arXiv:2503.05646},
  year={2025}
}

@inproceedings{moghani2024sufia,
  title={{SuFIA}: language-guided augmented dexterity for robotic surgical assistants},
  author={Moghani, Masoud and others},
  booktitle={IROS},
  pages={6969--6976},
  year={2024},
  organization={IEEE}
}

@article{moghani2025sufia,
  title={{SuFIA-BC}: Generating High Quality Demonstration Data for Visuomotor Policy Learning in Surgical Subtasks},
  author={Moghani, Masoud and others},
  journal={arXiv:2504.14857},
  year={2025}
}

@article{bjorck2025gr00t,
  title={{GR00T} n1: An open foundation model for generalist humanoid robots},
  author={Bjorck, Johan and others},
  journal={arXiv:2503.14734},
  year={2025}
}

@article{park2025modality,
  title={Modality-Augmented Fine-Tuning of Foundation Robot Policies for Cross-Embodiment Manipulation on GR1 and G1},
  author={Park, Junsung and others},
  journal={arXiv:2512.01358},
  year={2025}
}

@misc{IsaacLabEvalTasks,
  title        = {{IsaacLabEvalTasks}: Benchmarking {GR00T} N1 Policy in Isaac Lab},
  author       = {{NVIDIA}},
  year         = {2025},
  howpublished = {\url{https://github.com/isaac-sim/IsaacLabEvalTasks}},
}

@article{agarwal2025cosmos,
  title={Cosmos world foundation model platform for physical {AI}},
  author={Agarwal, Niket and others},
  journal={arXiv:2501},
  year={2025}
}

@article{loh2024adavln,
  title={{AdaVLN}: Towards Visual Language Navigation in Continuous Indoor Environments with Moving Humans},
  author={Loh, Dillon and others},
  journal={arXiv:2411.18539},
  year={2024}
}

@article{huang2024rekep,
  title={{ReKep}: Spatio-temporal reasoning of relational keypoint constraints for robotic manipulation},
  author={Huang, Wenlong and others},
  journal={arXiv:2409.01652},
  year={2024}
}

@inproceedings{bonyani2025embodied,
  author    = {Mahdi Bonyani and others},
  title     = {Embodied {AI} in Unstructured 3D Spaces: Fusing Mid and Long-Range Sensing for Instruction-Aware Construction Robotics},
  year      = {2025},
}

@article{nguyen2024tacex,
  title={{TacEx}: GelSight Tactile Simulation in Isaac Sim--Combining Soft-Body and Visuotactile Simulators},
  author={Nguyen, Duc Huy and others},
  journal={arXiv:2411.04776},
  year={2024}
}

@article{liu2025pr2,
  title={{PR2}: A Physics-and Photo-Realistic Humanoid Testbed With Pilot Study in Competition},
  author={Liu, Hangxin and others},
  journal={Journal of Field Robotics},
  year={2025},
}

@article{tao2024maniskill3,
  title={Maniskill3: Gpu parallelized robotics simulation and rendering for generalizable embodied {AI}},
  author={Tao, Stone and others},
  journal={arXiv:2410.00425},
  year={2024}
}

@article{ren2024infiniteworld,
  title={{InfiniteWorld}: A unified scalable simulation framework for general visual-language robot interaction},
  author={Ren, Pengzhen and others},
  journal={arXiv:2412.05789},
  year={2024}
}

@article{zhang2025agentworld,
  title={{AgentWorld}: An interactive simulation platform for scene construction and mobile robotic manipulation},
  author={Zhang, Yizheng and others},
  journal={arXiv:2508.07770},
  year={2025}
}

@article{tai2025realmirror,
  title={{RealMirror}: A Comprehensive, Open-Source Vision-Language-Action Platform for Embodied AI},
  author={Tai, Cong and others},
  journal={arXiv:2509.14687},
  year={2025}
}

@article{kachaev2025mind,
  title={Mind and motion aligned: a joint evaluation {IsaacSim} benchmark for task planning and low-level policies in mobile manipulation},
  author={Kachaev, Nikita and others},
  journal={arXiv:2508.15663},
  year={2025}
}

@inproceedings{zhou2024towards,
  title={{Towards building AI-CPS with NVIDIA Isaac Sim: An industrial benchmark and case study for robotics manipulation}},
  author={Zhou, Zhehua and others},
  booktitle={ICSE},
  pages={263--274},
  year={2024}
}

@article{lin2025vlnverse,
  title={{VLNVerse}: A Benchmark for Vision-Language Navigation with Versatile, Embodied, Realistic Simulation and Evaluation},
  author={Lin, Sihao and others},
  journal={arXiv:2512.19021},
  year={2025}
}

@article{mao2025robot,
  title={Robot Learning from a Physical World Model},
  author={Mao, Jiageng and others},
  journal={arXiv:2511.07416},
  year={2025}
}

@article{kang2024far,
  title={How far is video generation from world model: {A physical law perspective}},
  author={Kang, Bingyi and others},
  journal={arXiv:2411.02385},
  year={2024}
}

@article{ahmed2024systemic,
  title={A systemic survey of the {Omniverse} platform and its applications in data generation, simulation and metaverse},
  author={Ahmed, Naveed and others},
  journal={Frontiers in Computer Science},
  volume={6},
  pages={1423129},
  year={2024},
}

@inproceedings{cordero2025dsr,
  title={{DSR} Framework: A Hybrid Approach for Accelerated Training of {AI} Models in Mobile Robotics},
  author={Cordero Pareja and others},
  booktitle={WEA},
  pages={150--161},
  year={2025},
  organization={Springer}
}

@article{zafra2025survey,
  title={Survey of Simulators for Deformable Objects in Robotics},
  author={Zafra-Navarro, Alberto and others},
  journal={Jornadas de Automática},
  number={46},
  year={2025}
}

@article{flores2025ros,
  title={{ROS}-Compatible Robotics Simulators for {Industry 4.0 and Industry 5.0}: A Systematic Review of Trends and Technologies},
  author={Flores Gonzalez, Jose M and others},
  journal={Applied Sciences},
  volume={15},
  number={15},
  pages={8637},
  year={2025},
}

@article{kargar2024emerging,
  title={Emerging trends in realistic robotic simulations: A comprehensive systematic literature review},
  author={Kargar, Seyed Mohamad and others},
  journal={IEEE Access},
  volume={12},
  pages={191264--191287},
  year={2024},
}

@article{gao2026reinforcementlearningfollowtheleaderrobotic,
      title={Reinforcement Learning for Follow-the-Leader Robotic Endoscopic Navigation via Synthetic Data}, 
      author={Sicong Gao and others},
      year={2026},
      journal={arXiv:2601.02798},
}

\end{document}